\documentclass[lettersize,journal]{IEEEtran}
\usepackage{amsmath,amsfonts}
\usepackage{algorithmic}
\usepackage{algorithm}
\usepackage{array}
\usepackage[caption=false,font=normalsize,labelfont=sf,textfont=sf]{subfig}
\usepackage{textcomp}
\usepackage{stfloats}
\usepackage{url}
\usepackage{verbatim}
\usepackage{graphicx}
\usepackage{cite}

\usepackage{tikz}
\usepackage{enumitem}
\usepackage{multicol}
\usepackage{amssymb}
\usepackage{multirow}
\usepackage{marvosym}
\usepackage[table]{xcolor}
\usepackage{booktabs} 
\begin{document}

\title{LSTM-MAS: A Long Short-Term Memory Inspired Multi-Agent System for Long-Context Understanding}

\author{
	\IEEEauthorblockN{
		Yichen Jiang\IEEEauthorrefmark{1}\IEEEauthorrefmark{2}, 
		Jiakang Yuan\IEEEauthorrefmark{1}\IEEEauthorrefmark{2}, 
		Chongjun Tu\IEEEauthorrefmark{1} 
		Peng Ye\IEEEauthorrefmark{2}\IEEEauthorrefmark{4}\textsuperscript{(\Letter)}
        and Tao Chen\IEEEauthorrefmark{1}\IEEEauthorrefmark{3}}\textsuperscript{(\Letter)} \\
	\IEEEauthorblockA{\IEEEauthorrefmark{1}Fudan University, Shanghai, China}
	\IEEEauthorblockA{\IEEEauthorrefmark{2}Shanghai Artificial Intelligence Laboratory, Shanghai, China}
	\IEEEauthorblockA{\IEEEauthorrefmark{3}Shanghai Innovation Institute, Shanghai, China} 
	\IEEEauthorblockA{\IEEEauthorrefmark{4}The Chinese University of Hong Kong, Hong Kong, China}
\thanks{The copyright belongs to \textit{IEEE Transactions on Knowledge and Data Engineering}.}
\thanks{\textsuperscript{(\Letter)}Corresponding author: Peng Ye, Tao Chen}
}

\markboth{LSTM-MAS: A Long Short-Term Memory Inspired Multi-Agent System for Long-Context Understanding}%
{Shell \MakeLowercase{\textit{et al.}}: A Sample Article Using IEEEtran.cls for IEEE Journals}

\IEEEpubid{0000--0000/00\$00.00~\copyright~2025 IEEE}

\maketitle

\begin{abstract}
Effectively processing long contexts remains a fundamental yet unsolved challenge for large language models (LLMs).
Existing single-LLM-based methods primarily reduce the context window or optimize the attention mechanism, 
but they often encounter additional computational costs or constrained expanded context length.
While multi-agent-based frameworks can mitigate these limitations,
they remain susceptible to the accumulation of errors and the propagation of hallucinations.
In this work, we draw inspiration from the Long Short-Term Memory (LSTM) architecture to design a Multi-Agent System called LSTM-MAS, emulating LSTM’s hierarchical information flow and gated memory mechanisms for long-context understanding.
Specifically, LSTM-MAS organizes agents in a chained architecture, where each node comprises a worker agent for segment-level comprehension, a filter agent for redundancy reduction, a judge agent for continuous error detection, 
and a manager agent for globally regulates information propagation and retention, analogous to LSTM and its input gate, forget gate, constant error carousel unit, and output gate.
These novel designs enable controlled information transfer and selective long-term dependency modeling across textual segments, which can effectively avoid error accumulation and hallucination propagation.
We conducted an extensive evaluation of our method. Compared with the previous best multi-agent approach, CoA, our model achieves improvements of \textbf{97.97\%}, \textbf{65.75\%}, \textbf{122.19\%}, \textbf{39.61\%} and \textbf{10.80\%} on Narrative QA, Qasper, HotpotQA, 2WikiMQA and MuSiQue, respectively.
\end{abstract}

\begin{IEEEkeywords}
Long-Context Understanding, Large Language Models, Multi-Agent System, Memory
\end{IEEEkeywords}

\section{Introduction}
\IEEEPARstart{L}arge Language Models (LLMs)~\cite{achiam2023gpt, bai2023qwentechnicalreport, cai2024internlm2, glm2024chatglm, grattafiori2024llama3herdmodels, longchat2023, minimax2025minimax01scalingfoundationmodels, zheng2023judging} have become a cornerstone of modern artificial intelligence, offering powerful capabilities through extensive knowledge and remarkable reasoning abilities. 
With the rapid development of large language models (LLMs), there is a growing expectation that such models can handle increasingly complex tasks while supporting substantially extended context lengths. This includes tasks like long-context question-and-answer, long-context summary, and few-shot learning~\cite{yang2018hotpotqa, ho20202WikiMQA, trivedi2022musique, zhong2021qmsum, huang2021efficient, zhang2021exploratory, 10508813, 10547346, 11051380, 9528998}.
These tasks require LLMs to gain a deep understanding of both the details and logical relationships within long contexts.
However, the transformer architectures of LLMs face a critical bottleneck: the attention mechanism scales quadratically with context length, leading to rapidly increasing computational and memory costs, which makes naive context scaling impractical for real-world deployment.

Existing methods for long-context understanding can be broadly classified into single-LLM-based and multi-agent-based methods.
The former category aims to enhance the long-context capability of a single LLM by either reducing the input length~\cite{team2023gemini, achiam2023gpt} (\textit{e.g.}, RAG~\cite{lewis2020retrieval}) or extending the context window (\textit{e.g.}, long-context fine-tuning)~\cite{chen2023extending, su2024roformer, munkhdalai2024leave, 10937248}.
While effective to some extent, these approaches either incur additional training costs or allow only constrained expansion of the context length.
The latter category leverages a multi-agent framework to tackle long-context tasks, often employing segmented processing to preserve high performance on short-context inputs while mitigating the exponential computational complexity.
LONGAGENT~\cite{zhao2024longagent} divides a long document into multiple segments, each processed by different agents, followed by multi-round aggregation to produce a unified output. Chain of Agents (CoA)~\cite{zhang2024chain} instead adopts a sequential collaboration strategy, where each agent processes its assigned segment conditioned on the output of the previous one. 
However, both approaches struggle to maintain global coherence due to the absence of explicit mechanisms for long-range dependency modeling and memory sharing, often resulting in information loss, accumulated errors, and hallucinated reasoning.
\IEEEpubidadjcol 

\begin{figure*}[!t]
  \centering
  \includegraphics[width=\linewidth]{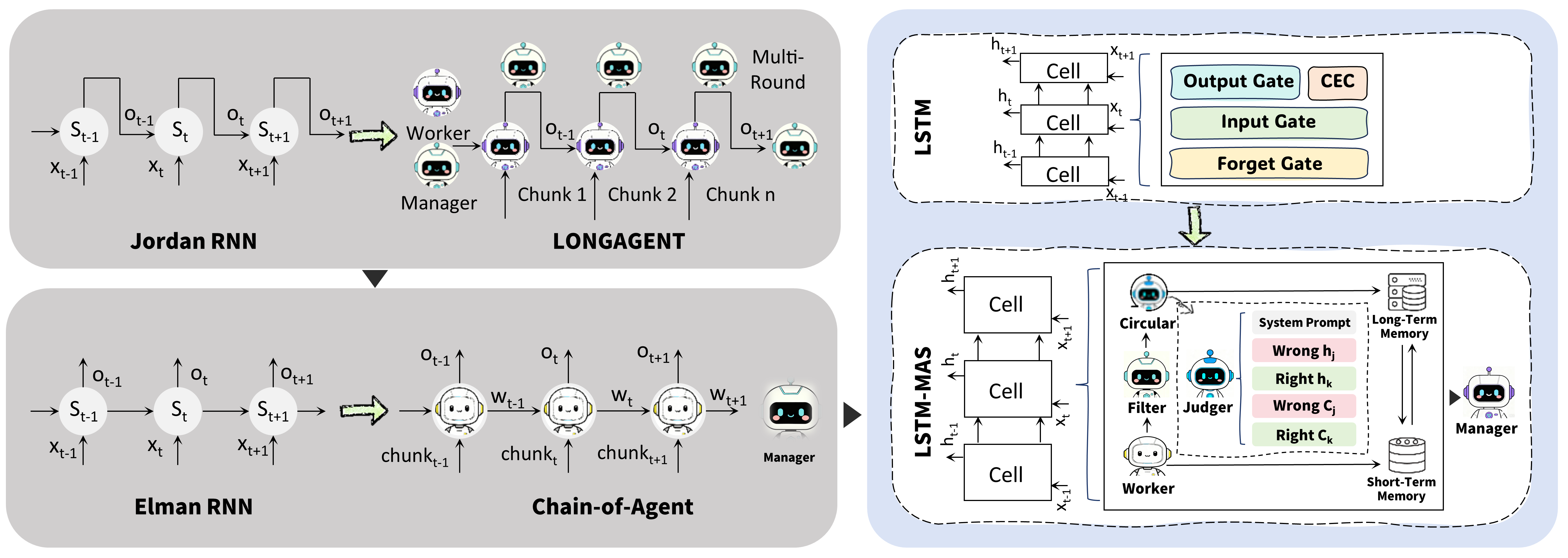}
  \caption{
  Analogical comparison between three multi-agent systems (MASs) and traditional neural network structures. Specifically, the upper left panel draws an analogy between LONGAGENT~\cite{zhao2024longagent} and the Jordan RNN~\cite{jordan1986attractor}; the lower left panel analogizes CoA~\cite{zhang2024chain} with the Elman RNN~\cite{elman1990finding}; and the right panel illustrates the analogical analysis between the proposed framework LSTM-MAS and the standard LSTM.
  }
  \label{fig:stru_comparison}
\end{figure*}

The rapid evolution of artificial intelligence has progressed from designing operators (\textit{e.g.}, convolutions), to designing deep models (\textit{e.g.}, Transformer), and more recently, to designing AI systems (\textit{e.g.}, multi-agent systems).
Learning from this developmental trajectory, we revisit the design of multi-agent systems for long-context modeling from the perspective of deep model design, seeking to transfer the principles that enables deep models to succeed into the design of multi-agent systems.
From this viewpoint, as illustrated in Figure~\ref{fig:stru_comparison}, LONGAGENT~\cite{zhao2024longagent} functions analogously to a Jordan recurrent neural network~\cite{jordan1986attractor}, where the leader aggregates member outputs for global control, while CoA~\cite{zhang2024chain} resembles an Elman recurrent neural network~\cite{elman1990finding}, which maintains context through internal recurrence.
However, such RNN-like architectures inevitably inherit the classical limitations of RNNs, such as error accumulation and limited long-range dependency modeling.
Inspired by the success of LSTM~\cite{hochreiter1997long} that introduces gating mechanisms and memory cells to reduce errors and preserve long-range information, we investigate \textit{how the core design principles of LSTM can be reformulated for multi-agent systems, enabling robust and effective long-context understanding.}

In this paper, we propose \textbf{LSTM-MAS}, a training-free multi-agent framework inspired by LSTM, as shown in Figure~\ref{fig:stru_comparison}.
LSTM-MAS forms an implicit state flow, analogous to LSTM propagates hidden states across time steps, where multiple nodes are sequentially connected, each node processes a fragment of the input text, and its output becomes the input for the next node.
More specifically, the system consists of four functional agents. 
The \textbf{\textit{Worker Agent}} performs segment-level comprehension by processing the current text fragment and integrating previous contextual information to generate a hidden representation, analogous to the hidden state in LSTM. 
The \textbf{\textit{Filter Agent}} , acting similarly to the forget gate in LSTM, evaluates the relevance of the Worker’s output to remove redundant, irrelevant, or hallucinatory information. 
The \textit{\textbf{Judge agent}} corresponds to the CEC unit in LSTM. When inconsistencies are detected, it re-examines the original text and flags potential reasoning errors for subsequent correction.
Finally, the function of \textbf{\textit{Manager Agent}} is similar to output gate in LSTM. It conducts global reasoning and summarizes the final answer for output.
This design simulates the memory cells and gating mechanisms of LSTM, which can effectively prevent error accumulation and hallucination propagation.

\begin{figure*}[!t]
  \centering
  \includegraphics[width=\linewidth]{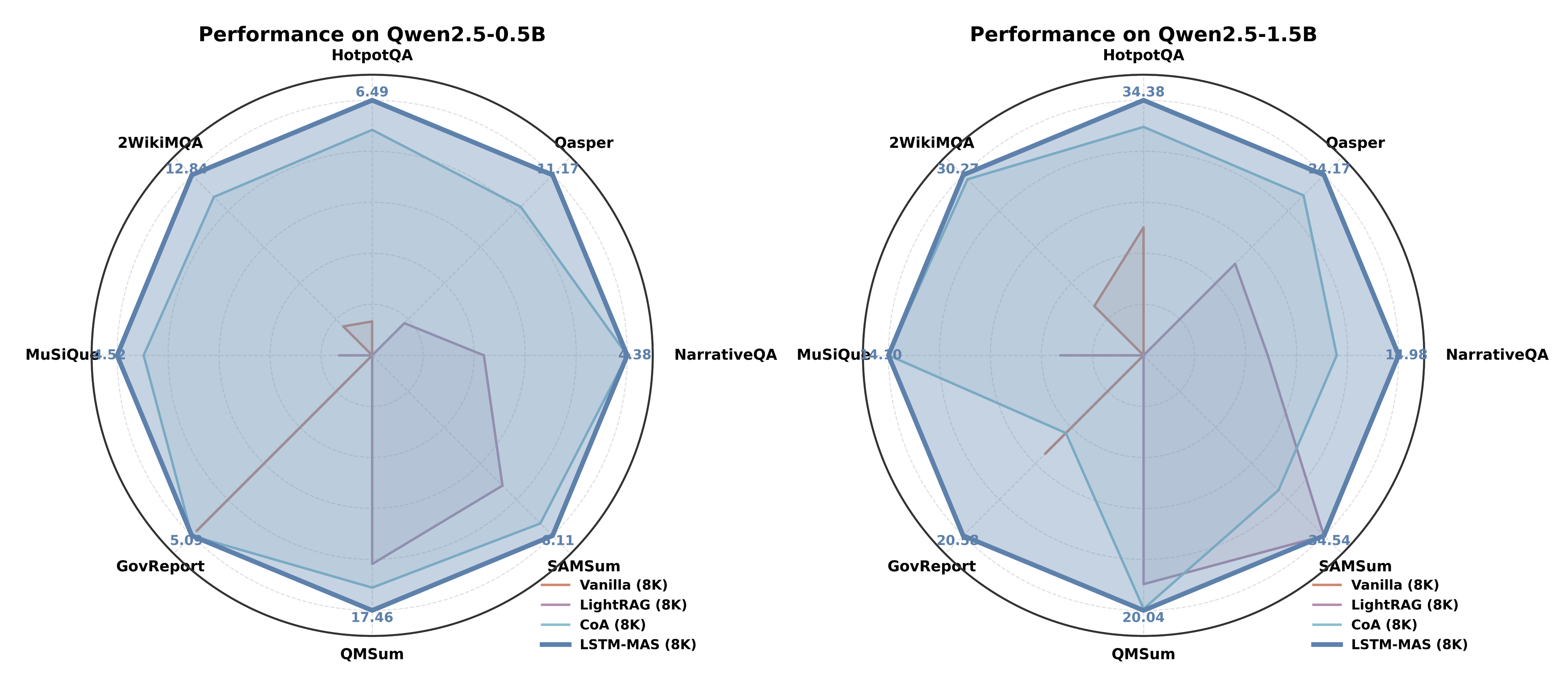}
  \caption{
  Test Results of Qwen2.5-0.5B (left) and Qwen2.5-1.5B (right) as base models on vanilla, LightRAG, CoA, and LSTM-MAS. LSTM-MAS achieved leading results on all eight datasets.
  }
  \label{fig:radar_fig}
\end{figure*}

We conducted extensive evaluations of our LSTM-MAS on eight datasets, covering a diverse range of task types, including single-document Q\&A, multi-document Q\&A, multi-hop Q\&A, long-context summarization, and few-shot learning.
As shown in Figure~\ref{fig:radar_fig}, LSTM-MAS performs as an all-round performer across all evaluation benchmarks, achieving substantial improvements over baselines such as Vanilla, LightRAG, and CoA.
In particular, compared with the current SOTA multi-agent baseline CoA, our method achieves improvements of \textit{\textbf{40.93\%}, \textbf{43.70\%},\textbf{ 121.57\%}} and \textit{\textbf{33.12\%}} on NarrativeQA, Qasper, HotpotQA, and MuSiQue, respectively.
These results demonstrate that LSTM-MAS effectively mitigates the problems of \textit{error accumulation} and \textit{hallucination propagation} that commonly arise in existing multi-agent frameworks, validating the soundness of leveraging the classical deep model design to guide the design of multi-agent systems.
Moreover, experiments across models of varying scales and architectures show consistent performance gains, indicating the strong generalization capability of LSTM-MAS, while its training-free nature further provides high flexibility.

\section{Related works}
\paragraph{Single-LLM-based Long-Context Understanding}
The first approach to utilizing a single large language model (LLM) for processing long-context tasks involves reducing the input length.
For instance, Retrieval-Augmented Generation (RAG)~\cite{lewis2020retrieval} relies on a retrieval module to select relevant fragments; InfLLM~\cite{xiao2024infllm} employs efficient contextual memory modules to identify pertinent information; Activation Beacon~\cite{zhang2024long} directly compresses the keys and values at each layer; and Landmark Attention~\cite{mohtashami2023landmark} trains the attention mechanism to utilize landmarks for selecting relevant blocks. Although this approach can alleviate the challenge of long contexts when combined with powerful retrievals~\cite{xiong2023examining, wang2022text, lin2023train, 8844864}, it still leads to the loss of intermediate information.
The second approach involves fine-tuning techniques that increase the context window of the model. For example, rotating position embeddings~\cite{su2024roformer} and T5 bias~\cite{raffel2020exploring} enable the extension of the context window to some degree by modifying the positional encoding strategy; ALiBi~\cite{press2021train} introduces a linear bias to the attention scores, replacing traditional positional encoding; and CLEX~\cite{chen2023clex} extends the context length via continuous positional embedding scaling. 
However, since these methods rely on additional training, they incur additional computational costs. Moreover, because the text length used during fine-tuning is fixed, the resulting models are constrained expanded context length.

\paragraph{Multi-Agent-based Long-Context Understanding}
Simultaneously, research has been conducted on multi-agent-based long-context tasks. For instance,
MEMWALKER~\cite{chen2023walking} addresses the issue of limited context windows by treating the LLM as an interactive agent, summarizing long contexts into a tree structure and conducting searches.
TRANSAGENTS~\cite{wu2024transagents} is employed for translating lengthy literary texts.
LONGAGENT~\cite{zhao2024longagent} assigns different roles to agents, with leaders responsible for understanding user intent, guiding team members to extract information from documents, and mitigating misunderstandings through communication among members. 
CoA~\cite{zhang2024chain} is a multi-agent system consisting of multiple worker agents that communicate sequentially to process different segmented parts of the text, followed by a manager agent that synthesizes these contributions into a coherent final output.
However, existing methods lack mechanisms to judge and filter intermediate outputs generated by agents, which can lead to accumulation of errors and the propagation of hallucinations.

\paragraph{Recurrent Neural Network and Their Variants~\cite{7508408}}
Recurrent Neural Networks (RNNs) and their variants have long been fundamental architectures for sequential data processing and temporal dependency modeling.
Jordan RNN~\cite{jordan1986attractor} directly feeds back the final output of the entire network to the input layer of the network after a delay. The output of a recurrent layer in Elman RNN~\cite{elman1990finding} takes the output of a recurrent layer, after a time delay, as part of the input for this layer at the next moment, and then simultaneously sends the output of the recurrent layer to the subsequent layers of the network, such as the final input layer.
The Long Short-Term Memory (LSTM~\cite{hochreiter1997long}) network introduced memory cells and gating mechanisms that selectively retain, update, and forget information, significantly enhancing stability in gradient propagation and temporal reasoning.
Inspired by the gating mechanism and memory cells of LSTM, we creatively designed a multi-agent system similar to the LSTM structure, using the gating and memory module to solve the problems of insufficient long-term dependency, error accumulation and hallucination propagation existing in the previous methods.

\begin{figure*}[!t]
  \centering
  \begin{tikzpicture}
    \node[inner sep=0pt, outer sep=0pt] (img) 
      {\includegraphics[width=\linewidth]{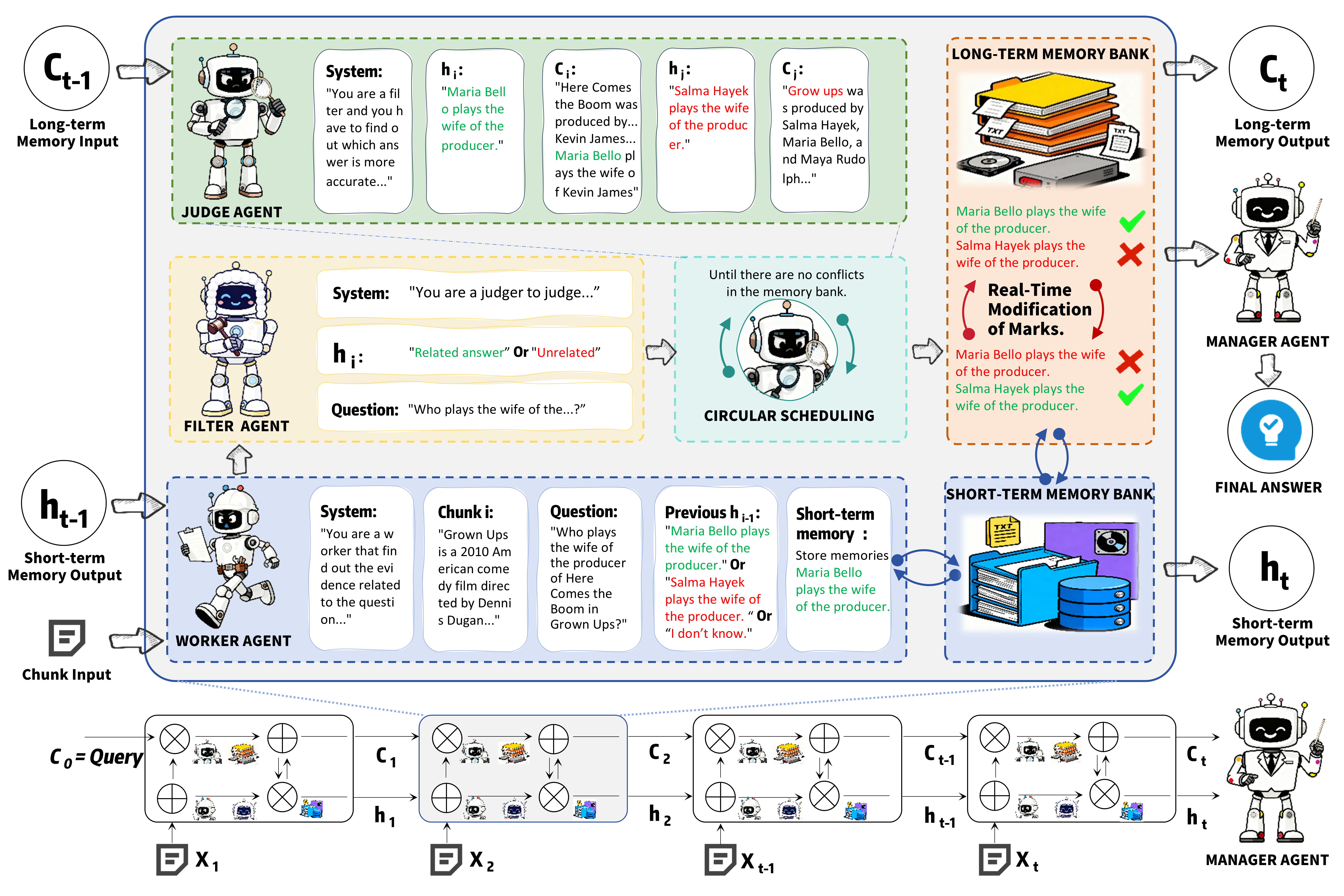}};
    \draw[dashed, thick, rounded corners=3pt] 
      (img.south west) rectangle (img.north east);
  \end{tikzpicture}
  \caption{Overview of LSTM-MAS, which is organized in a chain structure. Each node includes a worker agent, a filter agent and a judge agent. At the end of the chain is a manager agent responsible for answering questions based on the output of the entire chain.}
  \label{fig:overview}
\end{figure*}


\begin{algorithm}
\caption{Forward Pass of LSTM-MAS Node}
\label{alg:LSTM-MAS}
\renewcommand{\algorithmicrequire}{\textbf{Input:}}
\renewcommand{\algorithmicensure}{\textbf{Output:}}
    \begin{algorithmic}[1] 
    \REQUIRE System prompts $s_*$, source document $x$, query $q$, maximum retries $N_{max}$
    \ENSURE Final Answer to the query.
    \STATE Split $x$ into $l$ chunks $\{x_1, x_2, \cdots, x_l\}$ where $|x_i| \le k$
    \STATE Initialize long-term memory $C_0 \leftarrow \emptyset$, short-term state $h_0 \leftarrow \emptyset$
    \FOR {$i = 1$ \TO $l$}
        \STATE \textit{// 1. Worker Agent processes local and historical context}
        \STATE $h_i \leftarrow \text{LLM}_W(P_{worker}(s_W \oplus x_i \oplus q \oplus h_{i-1}))$
        
        \STATE \textit{// 2. Filter Agent decides retention}
        \STATE $o_{F,i} \leftarrow \text{LLM}_F(P_{filter}(s_F \oplus h_i \oplus q))$
        \STATE $\text{Decision}_i \leftarrow \mathbb{I}(\text{ExtractLabel}(o_{F,i}) \in \{\text{"related"}\})$
        
        \IF {$\text{Decision}_i = 1$}
            \STATE \textit{// 3. Judge Agent detects and resolves conflicts}
            \STATE $Conflicts \leftarrow \text{Detect}(h_i, C_{i-1})$ 
s            \IF {$Conflicts \neq \emptyset$}
                \FOR {\textbf{each} $h_j \in Conflicts$}
                    \STATE $iter \leftarrow 0$, $resolved \leftarrow \text{False}$
                    \WHILE {$iter < N_{max}$ \AND \textbf{not} $resolved$}
                        \STATE $J_{output} \leftarrow \text{LLM}_J(P_{judge}(s_J \oplus x_i \oplus h_i \oplus x_j \oplus h_j \oplus q))$
                        \STATE $resolved \leftarrow \text{Check\_Consensus}(J_{output})$
                        \STATE $iter \leftarrow iter + 1$
                    \ENDWHILE
                    
                    \IF {$resolved$}
                        \STATE $C_i \leftarrow (C_{i-1} \setminus \{h_j\}) \cup \{J_{output}\}$
                    \ELSE
                        \STATE Flag $h_i, h_j$ as "Unresolved" for Manager Agent
                        \STATE $C_i \leftarrow C_{i-1} \cup \{h_i, h_j\}$
                    \ENDIF
                \ENDFOR
            \ELSE
                \STATE $C_i \leftarrow C_{i-1} \cup \{h_i\}$
            \ENDIF
        \ELSE
            \STATE $C_i \leftarrow C_{i-1}$ \textit{// Forget mechanism}
        \ENDIF
    \ENDFOR
    \RETURN $\text{LLM}_M(P_{manager}(s_M \oplus C_l \oplus q))$
    \end{algorithmic}
\end{algorithm}

\section{Method}

\subsection{Preliminary}

\paragraph{Task Set}
We assume that a long context can be represented as $(x, y, q)$, where $x$ denotes the long context content, $y$ represents the text output, and $q$ is the optional text query. Let $x$ have a length of $n$ tokens and $y$ a length of $m$ tokens. Given a large language model (LLM) with a context window length of $k$ tokens ($k \ll n$), the objective is to generate the corresponding $y$ based on $q$ and $x$. To achieve this, we divide $x$ into $l$ blocks, such that $x = {x_1, x_2, \dots, x_l}$, and input these blocks sequentially into the system to simulate the sequence expansion of LSTM time steps.

\paragraph{Problem Set}
Existing multi-agent systems designed for long-context processing face two fundamental limitations: 
(1) the additional computational costs and constrained expanded context length, 
(2) the absence of the mechanisms of explicit mechanisms for long-range dependency modeling, which leads to the accumulation of errors and the propagation of hallucinations along the chain.

\paragraph{Overall Design}

Figure~\ref{fig:overview} illustrates the general framework of \textbf{LSTM-MAS}, whose structure is inspired by LSTM and adopts a chain-like connection form.   Each node functions as a “time step,” consisting of multiple collaborative agents that jointly accomplish the reasoning task for the current text block.   The input long context is first divided into several blocks and sequentially passed through the nodes in the chain.   While processing local content, each node maintains awareness of the semantics ahead, and its output serves as the input for the next node, forming an “implicit state flow” of information analogous to the hidden state propagation mechanism in LSTM.   At the end of the chain, the Manager Agent aggregates the outputs from all nodes to produce the final answer.

\subsection{Agent Design}

\paragraph{The Worker Agent}
The input to this agent consists of the current text segment and the "hidden state” from the previous node, while its output is the updated hidden state of the current node, while the hidden state constitutes short-term memory.
Our Worker Agent $W_i$ employs a dual-path contextual reasoning mechanism that explicitly integrates the understanding of the current input with dynamically maintained historical awareness, thereby addressing the issue of contextual incoherence across sequential segments.
Specifically, the Worker Agent processes the inputs through two coordinated pathways:
\textbf{(1) Current input processing}, which extracts salient information and contextual cues from the current segment $x_i$ under the guidance of the query $q$ and the working state $s_W$; and
\textbf{(2) Historical integration}, which retrieves and updates the prior hidden state $h_{i-1}$ to ensure long-range semantic consistency and continuity across nodes.

Formally, the Worker Agent updates its local historical awareness and produces a contextualized output. Unlike traditional recurrent networks that rely on continuous latent vectors, our agent operates on discrete token sequences. We define the updated contextual memory string $h_i$ as:
\begin{equation}
h_i = \text{LLM}_W(P_{worker}(s_W \oplus x_i \oplus q \oplus h_{i-1})), \label{eq:worker_update}
\end{equation}
where $\text{LLM}_W(\cdot)$ denotes the text generation function of the underlying large language model, and $\oplus$ represents the concatenation of contextual components. $P_{worker}$ is a structured prompt template that integrates the role-specific system instruction $s_W$, the current text segment $x_i$, the user query $q$, and the natural language reasoning trace from the previous step $h_{i-1}$.

This dual-path update compels the agent to simultaneously attend to the local context ($x_i$) and the historical reasoning state ($h_{i-1}$), enforcing a structured reasoning flow that mitigates short-sighted decision-making. Moreover, rather than maintaining a dense mathematical vector, $h_i$ is continually summarized and compressed through the LLM's semantic abstraction capabilities, preserving only essential relational and factual information. This ensures that $W_i$ maintains long-term contextual fidelity without overwhelming the context window---crucial for multi-hop or multi-document reasoning tasks where early evidence may influence later conclusions.
\paragraph{The Filter Agent}
The Filter Agent can determine the relevance of the output of the Worker Agent.
While the Worker Agent integrates local and historical information, unfiltered propagation of noisy or contradictory content can lead to cumulative reasoning errors.
Traditional systems such as CoA rely on static filtering thresholds, which tend to either retain excessive noise or mistakenly discard crucial evidence.
To address this limitation, the Filter Agent $F_i$ functions as a dynamic gating mechanism that adaptively regulates information flow between nodes, analogous to an information filter with task-specific awareness.
Specifically, $F_i$ operates along two synergistic dimensions:
\textbf{(1) Instruction-sensitive filtering.} The agent employs task-specific instructions $s_F$ to modulate its filtering strictness, adopting lenient thresholds for open-ended or creative tasks, and stricter control for factual reasoning and question answering.
\textbf{(2) Conflict-aware pruning.} $F_i$ simultaneously determines irrelevant information, model errors and hallucinations to ensure the robustness of the reasoning process.

Formally, given the reasoning trace $h_i$ from the Worker Agent and the user query $q$, the Filter Agent evaluates the relevance of the extracted information. Instead of relying on a continuous learned gating matrix, our training-free framework formulates this as a discrete prompt-based classification. We define the raw textual output of the Filter Agent as:
\begin{equation}
o_{F,i} = \text{LLM}_F(P_{filter}(s_F \oplus h_i \oplus q)), \label{eq:filter_out}
\end{equation}
where $\text{LLM}_F(\cdot)$ denotes the generation function of the Filter Agent, $P_{filter}$ is the specific prompt template, and $s_F$ dictates the task-specific strictness instructions.From this generated output, we extract a discrete relevance decision using an indicator function:
\begin{equation}
\text{Decision}
i = \mathbb{I}(\text{ExtractLabel}(o{F,i}) \in {\text{"related"}}), \label{eq:filter_decision}
\end{equation}
where $\mathbb{I}(\cdot)$ returns $1$ if the Worker Agent's output is deemed relevant and $0$ otherwise.Consequently, the aggregated long-term memory bank $C_i$, which tracks the preserved information across the chain, is updated conditionally:
\begin{equation}
C_i = \begin{cases}
C_{i-1} \cup \{ (x_i, h_i) \}, & \text{if } \text{Decision}_i = 1 \\  
C_{i-1}, & \text{otherwise}                                           
\end{cases} 
\label{eq:filter_context}
\end{equation}
Through this design, the Filter Agent dynamically balances information preservation and suppression. By explicitly acting as a discrete forget gate, it effectively filters out irrelevant noise and erroneous reasoning traces, thereby mitigating the risks of hallucination propagation and over-pruning without requiring any parameter updates.


\paragraph{The Judge Agent}
The Judge Agent ensures that the system possesses a feedback error correction mechanism, analogous to the Constant Error Carousel (CEC) in an LSTM. When contradictory reasoning is detected, the Judge Agent is triggered to reconcile the conflict. 

To address the definition of the conflict threshold in a training-free framework, we replace traditional numerical distance metrics with a semantic evaluation mechanism. The conflict detection function is defined as:
\begin{equation}
\text{Conflict}_{i,j} = \text{Detect}(h_i, h_j), \quad \exists h_j \in C_{i-1} \text{ where } j < i. \label{eq:conflict_detect}
\end{equation}
Here, $\text{Detect}(\cdot)$ utilizes an LLM prompt to compute a Boolean output, triggering a conflict only when explicit logical contradictions or mutually exclusive factual claims are identified between the current trace $h_i$ and the historical trace $h_j$.

Upon triggering, the Judge Agent retrieves the original text segments $x_i, x_j$ and the conflicting traces $h_i, h_j$. Instead of performing continuous vector multiplications, it utilizes the LLM's reasoning capabilities to derive a corrected, unified answer $J_{output}$ via a structured prompt:
\begin{equation}
J_{output} = \text{LLM}_J(P_{judge}(s_J \oplus x_i \oplus h_i \oplus x_j \oplus h_j \oplus q)). \label{eq:judge_resolve}
\end{equation}

Subsequently, the memory bank is discretely updated by replacing the flawed entries with the reconciled conclusion:
\begin{equation}
C_i = (C_{i-1} \setminus \{h_j\}) \cup \{J_{output}\}. \label{eq:memory_update}
\end{equation}

\textbf{Convergence Guarantee:} To ensure system resilience and prevent infinite loops when the Filter and Judge agents continuously disagree, we introduce a strict iteration constraint. The Judge Agent is permitted a maximum of $N_{max}$ iterations (empirically set to $N_{max} = 3$) to reach a consensus. If the conflict remains unresolved after $N_{max}$ attempts, the system enforces convergence by halting the loop, flagging both $h_i$ and $h_j$ as "unresolved," and forwarding them to the terminal Manager Agent. The Manager Agent, possessing the global context, makes the final arbitration. Through this mechanism, the accumulation of reasoning errors and the propagation of hallucinations are effectively suppressed while strictly guaranteeing logical halting.

\paragraph{The Manager Agent}
After dynamically generating multiple chain nodes based on the text length, the manager agent generates the final solution using the accumulated output at the end of the chain. Specifically, given the manager agent $M$, system prompt $s_M$, and query $q$, $M$ generates the final answer, $Answer$, based on the output $C_{\frac{n}{k}}$ at the end of the chain:
\begin{equation}
    Answer = \text{M}(s_M, C_{\frac{n}{k}}, q) \label{eq:manager_answer}
\end{equation}

The primary purpose of using a manager agent is to perform the final reasoning for the task. For instance, in a long context Q\&A task, the output at the end of the chain consists of a collection of "relevant evidence," and the manager agent performs the final reasoning for the last step based on these pieces of evidence.

The system prompts for the agents described above must be tailored to the specific task at hand. For instance, in Q\&A tasks, the worker agent is required to find evidence relevant to the problem; in text summarization tasks, the worker agent must summarize paragraphs and organize the logical relationships between the current and previous paragraphs; for other tasks, such as item counting, worker agents should be able to flexibly invoke relevant tools. The flexibility of the LSTM-MAS allows it to achieve strong performance across a variety of tasks.

\paragraph{Summary}
Overall, through the coordinated interaction of the Worker, Filter, and Judge Agents, LSTM-MAS maintains a continuous flow of long-context information throughout the reasoning chain, effectively mitigating the additional computational costs and realizing unconstrained expanded context length. Meanwhile, the Filter and Judge Agent enables handling of emerging errors in chain, preventing their accumulation of errors and the propagation of hallucinations.



\subsection{Mapping Relationship with LSTM}

The overall design of \textbf{LSTM-MAS} is inspired by the gating mechanism of Long Short-Term Memory (LSTM),
but extends it from the neuron-level to the agent-level.
As shown in Fig.~\ref{fig:overview}, the entire system forms a chain-structured reasoning process,
where each node corresponds to a “time step” in LSTM,
and the collaboration among its internal agents (Worker, Filter, and Judger) collectively realizes
the functions of input gate, forget gate, and CEC unit.
Meanwhile, the Manager Agent located at the end of the chain serves as an analog of the output gate,
aggregating all contextual representations to generate the final answer.
This section formally describes the mapping between the mathematical formulation of LSTM and that of LSTM-MAS.

\paragraph{Worker Agent $\leftrightarrow$ Input Gate}
In a standard LSTM, the input at time step $t$ is first projected to a hidden state:
\begin{equation}
\tilde{h}_t = \tanh(W_x x_t + W_h h_{t-1} + b),
\end{equation}
which fuses the current input $x_t$ and the previous hidden state $h_{t-1}$.
And as described earlier, the formula~\ref{eq:worker_update} of the \textit{Worker agent} also has a similar form.

\paragraph{Filter Agent $\leftrightarrow$ Forget Gate}
In LSTM, the forget and input gates are defined as:
\begin{equation}
f_t = \sigma(W_f [x_t;h_{t-1}] + b_f),
\end{equation}
which determine how much historical information is retained and how much new information is written.
Similarly, the formula~\ref{eq:filter_out} of \textit{Filter Agent} also decides which intermediate reasoning traces should be retained, suppressed, or corrected before flowing to the next node.

\paragraph{Judge Agent $\leftrightarrow$ Constant Error Carousel(CEC)}
In an LSTM, the cell state is updated as:
\begin{equation}
c_t = f_t \odot c_{t-1} + i_t \odot \tilde{h}_t.
\end{equation}
Similarly, the formula~\ref{eq:judge_resolve} of the Judge Agent indicates that the role of the Judge Agent is also to maintain the long-term consistency and stability of memory.

\paragraph{Manager Agent $\leftrightarrow$ Output Gate}
In LSTM, the output gate determines which part of the cell state is exposed to produce the hidden output. Evidently, the Manager Agent serves as the final reasoning controller, determining the ultimate content to be produced as output.

\section{Experiment}
\subsection{Experiment Setup}

\begin{table*}[!t]
  \caption{Datasets \& Metrics. Avg.Input is the average number of words in the dataset.}
  \label{datasets}
  \centering
  \resizebox{1.0\linewidth}{!}{
  \begin{tabular}{lcccccccc}
    \toprule
    ~ & \multicolumn{5}{c}{Q\&A} & \multicolumn{2}{c}{Summarization} & Few-shot Learning\\
    \cmidrule(r){2-6} \cmidrule(r){7-8} \cmidrule(r){9-9} 
     ~ & NarrativeQA & Qasper & HotpotQA & 2WikiMQA & MuSiQue & GovReport & QMSum & SAMSum  \\ 
    \midrule
    Avg.Input & 18,409 & 3,619 & 9,151 & 4,887 & 11,214 & 8,734 & 10,614 & 6,258  \\
    Eval metric & F1-score & F1-score & F1-score & F1-score & F1-score & Rouge-L & Rouge-L & Rouge-L   \\
    Query based & ${\surd}$ & ${\surd}$ & ${\surd}$ & ${\surd}$ & ${\surd}$ & ${\times}$ & ${\surd}$ & ${\surd}$ \\
    \bottomrule
  \end{tabular}
  }
\end{table*}

\paragraph{Datasets}
We tested LSTM-MAS on eight datasets in table~\ref{datasets}, including Q\&A tasks summarization tasks and few-shot learning tasks.
\begin{itemize}[leftmargin=15pt]
    \item \textbf{Q\&A tasks.} 
    We used five question-answering (QA) datasets for testing, following the evaluation settings of \textbf{LongBench}~\cite{bai2024longbench}. \textbf{NarrativeQA}~\cite{kovcisky2018narrativeqa} is a dataset containing human reading materials, where the system must read the entire book or movie script to answer questions related to the story. \textbf{Qasper}~\cite{dasigi2021dataset} is a dataset of multiple NLP papers that require searching for existing information throughout the entire text and providing answers written by a separate group of NLP practitioners. \textbf{HotpotQA}~\cite{yang2018hotpotqa} is a dataset designed for multi-hop questions based on Wikipedia, requiring the identification and reasoning across multiple supporting documents to generate answers. \textbf{2WikiMQA}~\cite{ho20202WikiMQA} is another multi-hop question-answering dataset that combines information from both Wikipedia and Wikidata. \textbf{MuSiQue}~\cite{trivedi2022musique} is a multi-hop problem dataset, but it contains more hops and additional distracting content compared to HotpotQA.
    \item \textbf{Summarization tasks.} 
    We used two summarization datasets for testing, also following the evaluation settings of LongBench. \textbf{GovReport}~\cite{huang2021efficient} is a long-document dataset composed of reports written by government research institutions, without task-specific input. \textbf{QMSum}~\cite{zhong2021qmsum} is a manually annotated benchmark dataset used for query-based, multi-domain conference summarization tasks. 
    \item \textbf{Few-shot Learning tasks.} 
    The \textbf{SAMSum}~\cite{gliwa2019samsum} dataset is a new dataset designed for generative dialogue summarization, which continues generating subsequent dialogues based on the few-shot dialogues mentioned earlier.
\end{itemize}

\paragraph{Eval metrics}
The question-answering (Q\&A) dataset is evaluated using the F1-score, while the summarization and SAMSum datasets are evaluated using Rouge-L scores.

\paragraph{Baselines} 
We primarily used three baseline frameworks for comparison. The first baseline is the vanilla model, which is the original base model that directly consumes tokens until its contextual window is fully utilized. The second baseline is LONGAGENT, which achieves the final result through segmenting and multiple rounds of summarization. The third baseline is CoA, which is currently the SOTA framework for multi-agent processing of long contexts and significantly improves evaluation metrics compared to RAG. During evaluation, we ensure that the text block division length for LSTM-MAS and CoA remains consistent. If a dataset does not present input issues, LSTM-MAS will automatically perform the summarization task.

\paragraph{Models}

We primarily used the \textbf{Qwen2.5-0.5B}~\cite{yang2024qwen2}, \textbf{Qwen2.5-1.5B}~\cite{yang2024qwen2}, \textbf{Qwen2.5-7B}~\cite{yang2024qwen2}, \textbf{Qwen2.5-14B}~\cite{yang2024qwen2}, \textbf{Qwen2.5-32B}~\cite{yang2024qwen2}, 
\textbf{Llama-3.2-3B-Instruct}~\cite{grattafiori2024llama3herdmodels}, \textbf{Llama-3.1-8B-Instruct}~\cite{grattafiori2024llama3herdmodels}, 
\textbf{Phi-3.5-mini-instruct}~\cite{abdin2024phi3technicalreporthighly}, 
and \textbf{Phi-3-small-8k-instruct}~\cite{abdin2024phi3technicalreporthighly}, 
models, deployed locally, as the base models for LSTM-MAS. To investigate the impact of different contextual window lengths on the performance of LSTM-MAS, we evaluated the performance of these models under 8K and 16K contextual window lengths, respectively. In the system, unless otherwise specified, the worker agent, judge agent, and manager agent use the same base model and share the same context window length.
During testing, if the base model utilizes a longer context window, the text blocks are divided into longer sections.

\subsection{Overall Results of LSTM-MAS}
\definecolor{myDarkGreen}{HTML}{006400}
\begin{table*}[!t]
  \caption{The performance of the LSTM-MAS method is compared against that of four baseline methods: Vanilla, LightRAG, LONGAGENT, and CoA.  Results with superior scores are highlighted to underscore their statistical significance.}
  \label{results}
  \centering
   \resizebox{1.0\linewidth}{!}{
  \begin{tabular}{lcccccccccc}
    \toprule
    ~ & ~ & \multicolumn{5}{c}{Q\&A} & \multicolumn{2}{c}{Summarization} & Few-shot Learning\\
    \cmidrule(r){3-7} \cmidrule(r){8-9} \cmidrule(r){10-10} 
     LLMs & Baselines & NarrativeQA & Qasper & HotpotQA & 2WikiMQA & MuSiQue & GovReport & QMSum & SAMSum & \textbf{Avg.}\\ 
     
    \midrule
    
    \multirow{5}{*}{Qwen2.5-0.5B} & Vanilla(8K) & 0.00 & 2.03 & 0.76 & 1.42 & 0.00 & 4.92 & 1.78 & 1.83 & 1.59\\
    & LightRAG & 1.56 & 3.09 & 0.26 & 0.13 & 0.35 & 2.81 & 13.98 & 4.68 & 3.36\\
    & LONGAGENT(8K) & 0.57 & 0.71 & 5.89 & 1.01 & 0.79 & 3.12 & 5.24 & 1.02 & 2.29\\
    & CoA(8K) & 4.98 & 9.02 & 5.60 & 10.93 & 3.94 & \textbf{5.15} & 15.73 & 5.75 & 7.66\\
    \rowcolor{gray!20}
    & LSTM-MAS(8K) & \textbf{8.20} & \textbf{11.17} & \textbf{10.10}& \textbf{17.72} & \textbf{4.52} & 5.09 & \textbf{17.46} & \textbf{6.11} & \textcolor{myDarkGreen}{\textbf{10.05}}\\
    
    \addlinespace
    
    \multirow{5}{*}{Qwen2.5-1.5B} & Vanilla(8K) & 0.00 & 3.65 & 18.79 & 13.02 & 0.50 & 9.65 & 3.27 & 11.19 & 7.51\\
    & LightRAG & 6.09 & 12.44 & 7.38 & 8.78 & 3.86 & 2.80 & 17.91 & \textbf{34.54} & 12.92\\
    & LONGAGENT(8K) & 1.91 & 9.44 & 6.31 & 9.46 & 3.83 & 1.82 & 12.25 & 0.04 & 5.63\\
    & CoA(8K) & 10.58 & 21.31 & 30.9 & 29.63 & \textbf{17.76} & 7.18 & 19.91 & 27.43 & 20.59\\
    \rowcolor{gray!20}
    & LSTM-MAS(8K) & \textbf{14.98} & \textbf{24.17} & \textbf{34.38} & \textbf{30.27} & 14.10 & \textbf{20.58} & \textbf{20.04} & 27.39 & \textcolor{myDarkGreen}{\textbf{23.24}}\\
    
    \addlinespace
    
    \multirow{4}{*}{Qwen2.5-7B} & Vanilla(8K) & 1.26 & 9.56 & 1.15 & 4.43 & 10.73 & 12.75 & 9.62 & 15.32 & 8.10\\
    & LONGAGENT(8K) & 7.07 & 8.09 & 17.11 & 8.01 & 8.18 & 0.64 & 9.55 & 0.13 & 7.35\\
    & CoA(8K) & 10.36 & 22.86 & 19.24 & 22.52 & 13.89 & \textbf{26.32} & 12.84 & 33.91 & 20.24\\
    \rowcolor{gray!20}
    & LSTM-MAS(8K) & \textbf{20.51} & \textbf{37.89} & \textbf{42.75} & \textbf{31.44} & \textbf{15.39} & 22.85 & \textbf{20.50} & \textbf{36.00} & \textcolor{myDarkGreen}{\textbf{28.42}}\\
    
    \addlinespace
    
    \multirow{4}{*}{Qwen2.5-14B} & Vanilla(8K) & 6.14 & 27.52 & 23.53 & 8.02 & 14.38 & 7.24 & 13.67 & 15.27 & 14.47\\
    & LONGAGENT(8K) & 10.56 & 12.51 & 24.91 & 12.35 & 12.43 & 0.33 & 11.83 & 1.54 & 10.81\\
    & CoA(8K) & 9.77 & 16.69 & 14.08 & 13.98 & 6.18 & \textbf{16.73} & 20.27 & 30.22 & 15.99\\
    \rowcolor{gray!20}
    & LSTM-MAS(8K) & \textbf{23.87} & \textbf{37.03} & \textbf{46.07} & \textbf{39.09} & \textbf{18.34} & 15.86 & \textbf{21.37} & \textbf{32.96} & \textcolor{myDarkGreen}{\textbf{29.32}}\\
    
    \addlinespace
    
    \multirow{3}{*}{Qwen2.5-32B} & Vanilla(8K) & 1.87 & 10.91 & 6.26 & 10.45 & 1.33 & 16.77 & 15.38 & 36.78 & 12.47\\
    & LONGAGENT(8K) & 5.07 & 15.08 & 15.16 & 7.09 & 8.67 & 8.89 & 10.34 & 1.60 & 8.99\\
    & CoA(8K) & 15.03 & 16.91 & 12.33 & 19.86 & 12.19 & \textbf{28.22} & 19.53 & 26.51 & 18.82\\
    \rowcolor{gray!20}
    & LSTM-MAS(8K) & \textbf{21.96} & \textbf{31.74} & \textbf{41.54} & \textbf{36.32} & \textbf{16.28} & 18.44 & \textbf{20.44} & \textbf{36.82} & \textcolor{myDarkGreen}{\textbf{27.94}}\\
    
    \midrule
    
    \multirow{5}{*}{Llama-3.2-3B-Instruct} & Vanilla(8K) & 0.00 & 21.23 & 3.61 & 11.17 & 0.13 & 2.51 & 3.06 & 5.90 & 5.95\\
    & LightRAG & 5.30 & 11.41 & 6.46 & 7.27 & 3.64 & 2.80 & 17.82 & 0.37 & 7.39\\
    & LONGAGENT(8K) & 4.82 & 4.01 & 16.80 & 6.56 & 8.27 & 0.67 & 8.47 & 1.19 & 6.35\\
    & CoA(8K) & 11.20 & \textbf{28.34} & 39.28 & 29.66 & 15.17 & 13.25 & \textbf{17.86} & 7.04 & 20.22\\
    \rowcolor{gray!20}
    & LSTM-MAS(8K) & \textbf{17.72} & 21.81 & \textbf{42.4} & \textbf{33.24} & \textbf{21.4} & \textbf{14.19} & 16.53 & \textbf{9.01} & \textcolor{myDarkGreen}{\textbf{22.04}}\\
    
    \addlinespace
    
    \multirow{5}{*}{Llama-3.1-8B-Instruct} & Vanilla(8K) & 0.00 & 24.78 & 4.96 & 12.46 & 0.45 & 3.35 & 3.37 & 14.88 & 8.03\\
    & LightRAG & 5.23 & 7.42 & 5.76 & 6.13 & 4.27 & 2.81 & 19.76 & 21.91 & 10.00\\
    & LONGAGENT(8K) & 11.87 & 12.46 & 24.27 & 12.5 & 13.43 & 1.01 & 12.9 & 2.04 & 11.31\\
    & CoA(8K) & 9.02 & \textbf{35.68} & 38.92 & \textbf{36.6} & 18.09 & \textbf{19.51} & \textbf{16.87} & 14.97 & 23.71\\
    \rowcolor{gray!20}
    & LSTM-MAS(8K) & \textbf{14.83} & 32.24 & \textbf{50.19} & 35.81 & \textbf{24.40} & 16.33 & 15.83 & \textbf{25.99} & \textcolor{myDarkGreen}{\textbf{26.95}}\\
    
    
    
    \midrule
    
    \multirow{4}{*}{Phi-3.5-mini-instruct} & Vanilla(8K) & 0.00 & 10.12 & 2.18 & 7.39 & 0.00 & 8.91 & 2.12 & 10.09 & 5.10\\
    & LONGAGENT(8K) & 4.27 & 9.62 & 16.19 & 10.23 & 10.19 & 0.25 & 12.80 & 12.75 & 9.54\\
    & CoA(8K) & 5.46 & 29.89 & 28.35 & 7.72 & 7.98 & \textbf{22.64} & 19.02 & 15.18 & 17.03\\
    \rowcolor{gray!20}
    & LSTM-MAS(8K) & \textbf{13.01} & \textbf{31.04} & \textbf{33.14} & \textbf{17.98} & \textbf{17.63} & 20.46 & \textbf{20.71} & \textbf{16.71} & \textcolor{myDarkGreen}{\textbf{21.34}}\\
    
    
    
    \addlinespace

    \multirow{3}{*}{Phi-3-small-8k-instruct} & Vanilla(8K) & 0.00 & 19.00 & 3.73 & 11.74 & 0.11 & 9.25 & 3.13 & 15.22 & 7.77\\
    & LONGAGENT(8K) & 2.71 & 9.28 & 20.76 & 8.55 & 11.15 & 0.23 & 7.16 & 0.23 & 7.51\\
    & CoA(8K) & 13.90 & 30.03 & 47.33 & \textbf{45.71} & 16.40 & \textbf{21.13} & \textbf{21.69} & \textbf{41.62} & 29.73\\
    \rowcolor{gray!20}
    & LSTM-MAS(8K) & \textbf{18.95} & \textbf{36.90} & \textbf{50.02} & 45.31 & \textbf{29.46} & 18.63 & 20.90 & 35.40 & \textcolor{myDarkGreen}{\textbf{31.95}}\\

    \bottomrule
  \end{tabular}
  }
\end{table*}

Table~\ref{results} presents the test results of LSTM-MAS across eleven large language models (LLMs).
\paragraph{Q\&A}
First, LSTM-MAS (8K) demonstrated significant improvements over the baseline models across five question-answering datasets. When utilizing Qwen2.5-7B as the base model, our framework achieved substantial, multi-fold performance gains across all evaluated datasets. Furthermore, the experiments reveal that LSTM-MAS operates unconstrained by the intrinsic context window limits of the base models, effectively bridging the capabilities gap between small and large-scale models. Notably, within the Qwen2.5 series (7B, 14B, and 32B), the performance disparity was drastically reduced from approximately 44\% in the base models to a mere \textbf{4.7\%}.
Compared to the state-of-the-art result of CoA(8K) for current multi-agent systems, LSTM-MAS achieved improvements of \textbf{97.97\%}, \textbf{65.75\%}, \textbf{122.19\%}, \textbf{39.61\%} and \textbf{10.80\%} on Narrative QA, Qasper, HotpotQA, 2WikiMQA and MuSiQue, respectively, using the same base model, Qwen2.5-7B.
We also compared LightRAG~\cite{guo2024lightrag} with our method, due to insufficient computing power, we only compared LightRAG on smaller models (The results of the larger models are presented in Appendix C). LightRAG performed reasonably well on few-shot tasks; however, its performance on other datasets did not reach the level achieved by our approach.

\paragraph{Summarization and Few-shot Learning}
LSTM-MAS demonstrates equal effectiveness in both summarization and few-shot learning tasks. When using \textbf{Qwen2.5-7B} as the base model, LSTM-MAS(8K) achieved improvements of \textbf{79.22\%}, \textbf{113.10\%} and \textbf{134.99\%} on the GovReport, QMSum and SAMSum datasets, respectively, compared to Vanilla(8K). When compared to CoA(8K), our system showed an improvement of \textbf{59.7\%} and \textbf{7.73\%} on QMSum and SAMSum.
Compared with LONGAGENT, our framework has achieved significant gains. LONGAGENT is almost completely unable to handle tasks such as summarization and few-shot learning, while our framework demonstrates excellent generalization.

Concurrently, we observed a slight performance drop compared to CoA on the GovReport dataset for certain models. A detailed comparison between GovReport and another summarization benchmark, QMSum, reveals that this minor discrepancy stems from two primary factors. First, the prompt guiding our framework's Manager Agent implicitly favors concise outputs, whereas the ground truth references in GovReport are considerably longer. This length mismatch inherently penalizes the ROUGE-L score calculation. Second, unlike QMSum, which provides explicit queries detailing what content to summarize, GovReport lacks specific instruction inputs. To strictly adhere to the LongBench experimental settings, we abstained from providing additional heuristic instructions for GovReport. In the absence of a clear guided query, this unconstrained setting occasionally caused our framework to deviate from the optimal summarization focus.

\paragraph{16K window length}
Table~\ref{results-16K} presents the results of LSTM-MAS(16K). When the context window is set to 16K, Vanilla’s performance is significantly improved; however, LSTM-MAS remains effective across most datasets. Narrative QA, HotpotQA, and 2WikiMQA continue to experience input length issues. Compared to LSTM-MAS(8K), LSTM-MAS(16K) shows improvements of \textbf{9.86\%}, a decrease of 6.85\%, and an increase of \textbf{7.48\%} on these three datasets, respectively. This proves that the ultimate capability of LSTM-MAS \textbf{does not} depend on the length of the context window of the underlying model.

\section{Analysis}

\begin{figure*}[!t]
  \centering
  \includegraphics[width=\linewidth]{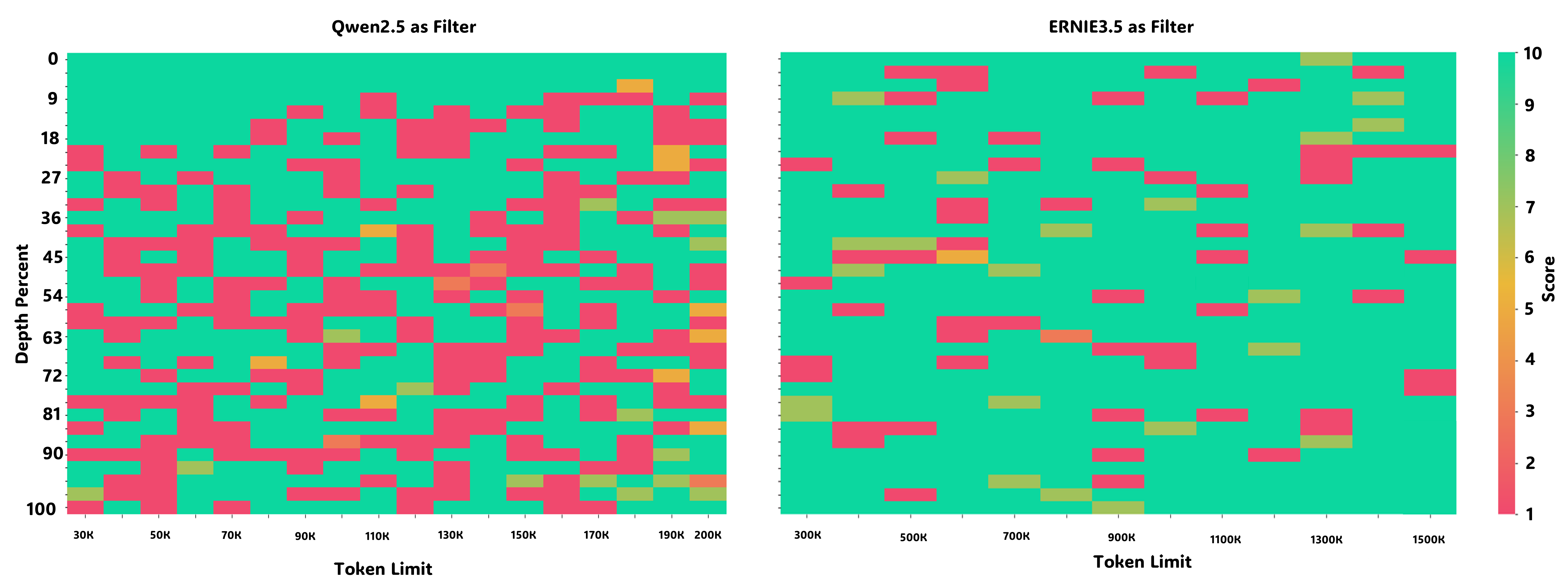}
  \caption{Needle-in-Haystack Test Results with Qwen2.5(left) and ERNIE 3.5~\cite{zhang2019ernie}(right) models as Filter. This indicates that the different series models are more suitable for the role of Filter agent and also highlights that heterogeneous models may cooperate effectively. Although this test could continue indefinitely, testing was stopped when the input length reached 1500K due to time constraints.}
  \label{fig:needle}
\end{figure*}

\begin{table}
  \caption{Comparison of the results of LSTM-MAS with Vanilla and CoA under 16K window (better results are indicated in bold).}
  \label{results-16K}
  \centering
  \setlength{\tabcolsep}{3pt}
  \resizebox{1.0\linewidth}{!}{
  \begin{tabular}{llccccc}
    \toprule
     LLMs & Baselines & NarrativeQA & Qasper & HotpotQA & 2WikiMQA & MuSiQue \\ 
    
    \midrule
    
    \multirow{3}{*}{Qwen2.5-7B} & Vanilla(16K) & 2.06 & 10.00 & 3.95 & 5.41 & 10.73\\
    & CoA(16K) & 11.67 & 22.52 & 16.33 & 15.15 & 8.50 \\
    & LSTM-MAS(16K) & \textbf{16.04} & \textbf{33.71} & \textbf{39.71} & \textbf{35.80} & \textbf{20.42}\\
    
    \midrule
    
    \multirow{3}{*}{Qwen2.5-14B} & Vanilla(16K) & 1.66 & 13.10 & 5.83 & 10.12 & 1.28\\
    & CoA(16K) & 10.57 & 20.76 & 13.08 & 15.84 & 6.79\\
    & LSTM-MAS(16K) & \textbf{24.28} & \textbf{36.87} & \textbf{49.07} & \textbf{41.03} & \textbf{21.56} \\
    \bottomrule
  \end{tabular}
  }
\end{table}

\begin{table}
  \caption{Ablation Experiment.}
  \label{ablation}
  \centering
  \setlength{\tabcolsep}{3pt}
  \resizebox{1.0\linewidth}{!}{
      \begin{tabular}{llcccccc}
        \toprule
         LLMs & Baselines & NarrativeQA & Qasper & HotpotQA & 2WikiMQA & MuSiQue & \textbf{Avg.}\\ 
        
        \midrule
        
        \multirow{3}{*}{Qwen2.5-7B} 
        & LSTM-MAS & 20.51 & 37.89 & 42.75 & 31.44 & 15.39 & \textbf{29.60}\\
        & -Judger & 14.2 & 35.86 & 20.8 & 25.98 & 12.65 & \textcolor{red}{-5.07}\\
        & -Filter & 13.49 & 30.81 & 35.15 & 29.76 & 14.24 & \textcolor{red}{-4.91}\\
        & -Manager & 9.02 & 20.05 & 10.86 & 10.94 & 5.29 & \textcolor{red}{-18.36}\\
        
        \midrule
        
        \multirow{3}{*}{Qwen2.5-14B} 
        & LSTM-MAS &  23.87 & 37.03 & 46.07 & 39.09 & 18.34 & \textbf{32.88}\\
        & -Judger & 23.84 & 36.16 & 44.87 & 33.17 & 16.71 & \textcolor{red}{-1.93}\\
        & -Filter & 12.20 & 21.85 & 35.26 & 31.19 & 13.33 & \textcolor{red}{-8.92}\\
        & -Manager & 9.93 & 21.83 & 13.43 & 12.10 & 6.89 & \textcolor{red}{-16.73}\\
        
        \bottomrule
      \end{tabular}
  }
\end{table}

\subsection{The Maximum Text Length That LSTM-MAS Can Handle}
In theory, LSTM-MAS can handle infinitely long contexts by constructing a sufficient number of agents that perform reasonable segmentation and configure the agents for processing, regardless of the input text’s length. However, in practice, it is still constrained by factors such as GPU memory size and disk capacity. To minimize the impact of these limitations, LSTM-MAS designs the text segment input and agent processing in a chained manner, so that at any given time, the GPU memory load is restricted to the encoding and inference of text blocks of length $k$. As for potential disk capacity constraints, given the large capacity of modern external storage devices, the likelihood of such limitations occurring is minimal. Even if issues arise, they can be addressed by swapping the most distant historical records. The design of this swapping strategy is system-independent and easily customizable. We conducted a "Needle in Haystack" test on LSTM-MAS, and the results are shown in Figure~\ref{fig:needle}. 
The maximum context length evaluated in our experiments reached 1500K tokens. Through the Needle-in-a-Haystack (NIAH) evaluation, we observed that employing a distinct base model for the Judge Agent significantly reduced the error rate, even at extended input lengths. We hypothesize that models sharing the same architecture and parameter scale exhibit highly similar hidden state distributions, making it inherently difficult for them to detect flaws in their own generated reasoning. This overlapping blind spot can be effectively mitigated by introducing a heterogeneous, larger-scale model as the judge.

\subsection{A More Reasonable Multi-Agent Cooperation Strategy Can Achieve Complex Reasoning in Long-Context tasks}
\begin{figure*}[!t]
  \centering
  \includegraphics[width=\linewidth]{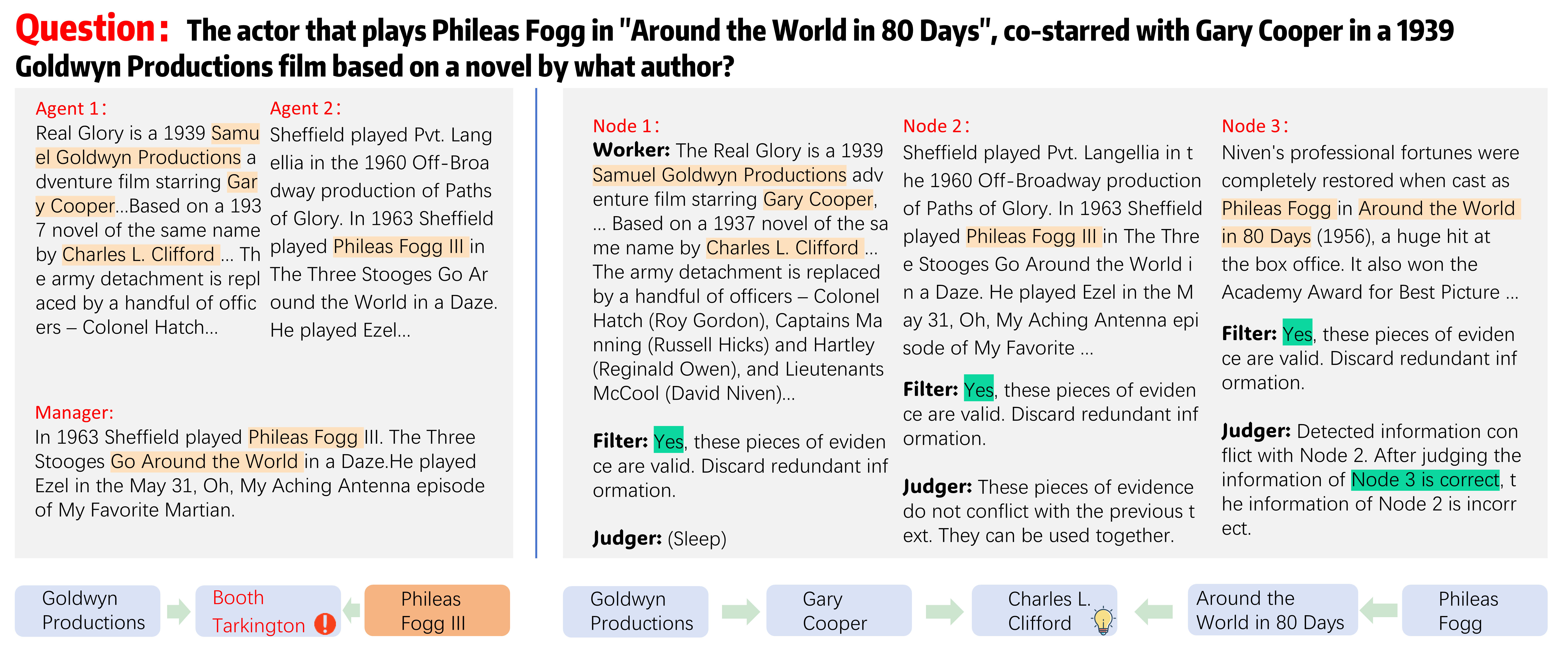}
  \caption{A case study of CoA(left) and LSTM-MAS(right). This demonstrates that the filter and judger in LSTM-MAS can effectively recover errors in the chain and improve the accuracy of the results.}
  \label{fig:case_study}
\end{figure*}

The effectiveness of the LSTM-MAS collaboration strategy is primarily reflected in the following aspects. First, the divide-and-conquer strategy proves effective for long-context processing. A single model is inherently limited by the context window length and is unable to process ultra long contexts. LSTM-MAS overcomes this limitation by dynamically constructing processing nodes. Even if the input text is excessively long, resulting in an extended chain, this scenario can be abstracted into a similar long-context problem that can be recursively addressed through divide-and-conquer methods. Second, the modular division of labor is highly rational, as illustrated in the processing procedure shown in Figure~\ref{fig:case_study}. The worker agent focuses on local information extraction, the filter agent evaluates the relevance of answers, the judge agent resolves conflicts, and the manager agent integrates new global information. Each agent is designed to focus on a single capability, thereby avoiding the "overload" of a single model. Finally, the redundancy and verification mechanism is essential. By comparing the answers of different worker agents, potential conflicts are identified and arbitrated, preventing accumulated errors and hallucination propagation.

\subsection{LSTM-MAS Can Narrow The Gap Between Small And Large Models In Long-Context Tasks Processing}
With a well-designed multi-agent cooperation strategy, the gap between small and large models in long-context tasks processing can be significantly reduced through system-level optimization and task decomposition. The core idea is to use collaborative strategies to compensate for the limitations of individual model capabilities. One approach is to simulate long-context tasks processing through divide-and-conquer strategies and memory transfer, effectively replacing expensive long-context attention mechanisms. Additionally, by leveraging multiple rounds of interaction between agents for cross-validation and conflict resolution, the issue of insufficient semantic understanding depth in smaller models is mitigated, leading to significant improvements in their effectiveness. When the base model has a large number of parameters and enhanced capabilities, shorter chains are typically employed, with more tasks assigned to each agent, thus reducing the number of agents and minimizing the hallucination problem in the interaction process. As shown in Table~\ref{ablation}, combining the capabilities of the filter and judge agent results in improved performance. The flexibility of LSTM-MAS allows for the easy customization of chain structures to accommodate different base models.

\subsection{Ablation Experiment}
Table~\ref{ablation} shows the results of the ablation study on various agents in the system. Since worker agents form the core of the LSTM-MAS framework, removing them would strip the system of its fundamental capabilities. Therefore, our ablation focuses only on the remaining three agent types. When the Filter Agent and Judge Agent are removed, the system’s performance drops significantly, confirming that their roles parallel the LSTM forget gate and CEC unit by ensuring proper memory transmission along the chain. Removing the Manager Agent likewise leads to a substantial performance decline, validating its necessity in the final reasoning stage. This also demonstrates that, analogous to the LSTM output gate, the Manager Agent effectively produces the final, meaningful memory output.
It can be observed that the influence of the Manager on the result is greater than that of the Filter and Judge. This is primarily because while the Filter and Judge focus on refining and maintaining the quality of intermediate memories, the Manager is indispensable for aggregating distributed pieces of evidence across the entire context to conduct the final global reasoning and synthesize the ultimate answer.

\begin{table*}
  \caption{Time consumption (Shorter time consumption is indicated in bold).}
  \label{time_table}
  \tiny
  \centering
  \resizebox{1.0\linewidth}{!}{
    \begin{tabular}{lccccccccc}
      \toprule
      LLMs & Baselines & NarrativeQA & Qasper & HotpotQA & 2WikiMQA & MuSiQue & QMSum & SAMSum & \textbf{Avg.} \\ 
      \midrule


      \multirow{2}{*}{Qwen2.5-1.5B} & LightRAG & \textbf{19.84} & 24.74 & 19.89 & 19.01 & 26.22 & 27.12 & 16.07 & 21.84\\
      & LSTM-MAS & 27.44 & \textbf{6.36} & \textbf{16.72} & \textbf{8.26} & \textbf{18.65} & \textbf{18.13} & \textbf{9.07} & \textcolor{myDarkGreen}{\textbf{14.95}}\\

      \midrule

      \multirow{2}{*}{Llama-3.2-3B-Instruct} & LightRAG & 33.64 & 48.44 & 32.20 & 29.44 & 30.11 & 49.80 & 46.92 & 38.65\\
      & LSTM-MAS & \textbf{20.64} & \textbf{5.38} & \textbf{10.63} & \textbf{5.72} & \textbf{14.30} & \textbf{15.68} & \textbf{7.56} & \textcolor{myDarkGreen}{\textbf{11.42}}\\

      \addlinespace

      \multirow{2}{*}{Llama-3.1-8B-Instruct} & LightRAG & 67.20 & 94.72 & 61.88 & 51.12 & 63.61 & 85.01 & 57.14 & 68.67\\
      & LSTM-MAS & \textbf{36.30} & \textbf{10.25} & \textbf{15.32} & \textbf{11.25} & \textbf{30.05} & \textbf{29.68} & \textbf{14.13} & \textcolor{myDarkGreen}{\textbf{21.00}}\\

      \midrule

      \multirow{2}{*}{Phi-3-mini-4k-instruct} & LightRAG & \textbf{8.20} & 14.09 & 18.89 & 17.11 & 24.37 & \textbf{14.91} & 17.31 & 16.41\\
      & LSTM-MAS & 26.08 & \textbf{7.12} & \textbf{14.21} & \textbf{7.57} & \textbf{16.80} & 24.36 & \textbf{11.01} & \textcolor{myDarkGreen}{\textbf{15.31}}\\
      
      \bottomrule
    \end{tabular}%
  }
\end{table*}

\section{Conclusion}
This paper introduces LSTM-MAS, an innovative framework for processing long-context tasks via multi-agent collaboration. It overcomes the limitation of single-LLM context windows by dynamically partitioning and linking intelligent agents to collaboratively process arbitrarily long-context tasks. The LSTM-MAS framework requires no training and offers high flexibility, supporting various tasks and model sizes while optimizing efficiency through dynamic adjustments in chain length and role allocation. Experimental results demonstrate that LSTM-MAS achieves significant improvements across multiple tasks compared to both the Vanilla model and the current state-of-the-art framework, CoA.

\textbf{Limitations: }
Although partitioning reduces the workload of individual computations, the serial processing of chains may result in accumulated latency, particularly when the text is extremely long, leading to decreased time efficiency of the framework. While the natural language-based agent communication mechanism offers high interpretability, it remains uncertain whether this is the most optimal communication method for agents. Future research should explore the possibility of enhancing communication efficiency between agents through fine-tuning and other approaches. Additionally, addressing issues such as limiting hallucinations and developing more sophisticated communication methods will further improve the performance of multi-agent systems in processing long-context tasks.

{\appendices
\section{Time Efficiency}
We conducted a comparative analysis of the time efficiency of our method against the LightRAG~\cite{guo2024lightrag} benchmark on the same dataset. The results indicate that the LLM-based approach achieves substantially lower time consumption compared to the retrieval-based method.
The results are presented in Table~\ref{time_table}.

Although the chained architecture of LSTM-MAS operates serially—which intuitively suggests higher latency—it achieves counter-intuitive speedups on specific long-document datasets. This phenomenon is primarily driven by the Filter Agent, which actively prunes a massive volume of irrelevant tokens early in the reasoning chain. Given that the standard self-attention mechanism in LLMs exhibits a quadratic computational complexity of $\mathcal{O}(N^2)$ with respect to the sequence length $N$, this dynamic token dropping yields outsized computational benefits. By drastically reducing the sequence length of the aggregated memory passed to subsequent nodes and the terminal Manager Agent, LSTM-MAS exponentially shrinks the size of the attention matrices required in later stages. Consequently, the computational time saved by avoiding dense $\mathcal{O}(N^2)$ calculations on full-length documents far outweighs the serial routing overhead, realizing an anomalous reduction in total inference time.

\begin{table}[b!]
  \caption{Prompt of agents in experiments.}
  \label{prompt}
  \centering
  \setlength{\tabcolsep}{3pt}
  \resizebox{0.77\linewidth}{!}{
  \begin{tabular}{ll}
    \toprule
    
    \multirow{3}{*}{Vanilla} 
    & Text: \{The whole text $C$.\}\\
    & Question: \{Question $q$.\}\\
    & Answer: \\
    
    \midrule
    
    
    \multirow{33}{*}{LSTM-MAS} 
    & \{Task Classification.\}\\
    & \textbf{Manager:}\\
    & System Prompt: \{$s_M$\} \\
    & **Example: **\\
    & **Question: **\\
    & The actor that plays Phileas Fogg in..., \\
    & co-starred with Gary Cooper in a 1939 \\
    & Goldwyn Productions film based on a \\
    & novel by what author? \\
    & **Your answer:** \\
    & Charles L. Clifford \\
    & Text chunk: \{chunk i: $x_i$.\} \\
    & Question: \{$q$.\} \\
    & Response:\{Final answer.\} \\
    & ~ \\
    & \textbf{Worker:} \\
    & System Prompt: \{$s_W$\} \\
    & Question: \{$q$.\} \\
    & Response: \{$h_i$.\} \\
    & ~ \\
    & \textbf{Filter:} \\
    & System Prompt: \{$s_F$\} \\
    & Question: \{$q$.\} \\
    & Answer: \{$F_{output}$.\} \\
    & Response: \{Related or Unrelated.\} \\
    & ~ \\
    & \textbf{Judge:} \\
    & System Prompt: \{$s_J$\} \\
    & Question: \{$q$.\} \\
    & history: \{$x_i, x_j$.\} \\
    & conflicts: \{$h_i, h_j$.\} \\
    & Response: \{$J_{output}$\} \\
    \bottomrule \\
  \end{tabular}
  }
\end{table}

\begin{table}[b!]
  \caption{Result of Larger Models on LightRAG.}
  \label{lrag}
  \centering
  \resizebox{0.77\linewidth}{!}{
  \begin{tabular}{ll}
    \toprule
    
    \multirow{3}{*}{output} 
    & [Item \{k\}] Query processed in x.xx seconds\\
    & response:  Sorry, I'm not able to provide an \\
    & \quad \quad \quad \quad answer to that question.[no-context]\\
    
    \bottomrule \\
  \end{tabular}
  }
\end{table}
\section{Experiments Details}
\paragraph{Agent selection and parameters for LSTM-MAS}
During the experiment, we divided text segments of different lengths based on the context window size of the basic model. When the context window was 16K, the length of the segmented text was 9000 tokens, and when the context window was 6000, the length of the segmented text was 5000 tokens.
\paragraph{Prompt engineering} The prompt designs for each agent in the experiment are shown in the table~\ref{prompt}.

\section{Hardware Information}
In the experiment, we used two NIVIDIA A800 80GB GPUs.
When running LightRAG, if the model is large, the log output will be as shown in Table~\ref{lrag}, and the final result has no statistical significance.

}

\bibliographystyle{IEEEtran}
\bibliography{reference}

\vfill

\end{document}